\documentclass[10pt,twocolumn,letterpaper]{article}

\usepackage{booktabs} 
\usepackage{cvpr}
\usepackage{times}
\usepackage{epsfig}
\usepackage{graphicx}
\usepackage{amsmath}
\usepackage{amssymb}
\usepackage{algorithm}
\usepackage{algorithmic}
\usepackage{subfigure}
\usepackage{url}
\usepackage{multirow}
\usepackage[usenames,dvipsnames,table]{xcolor}

\usepackage[breaklinks=true,bookmarks=false]{hyperref}

\cvprfinalcopy 

\def\cvprPaperID{****} 

\begin{document}

\title{ Matching-CNN Meets KNN:  Quasi-Parametric Human Parsing}

\author{
	Si~Liu$^{1,2}$\\
	\and
	Xiaodan~Liang$^{2,4}$\\
	\and
	Luoqi~Liu$^{2}$\\
	\and
	Xiaohui Shen$^{3}$\\
	\and
	Jianchao Yang$^{3}$\\
	\and
    Changsheng Xu$^{1,5}$\\
	\and
    Liang Lin$^{4}$\\
	\and
    Xiaochun Cao$^{1}$\\
	\and
	Shuicheng Yan$^{2}$\\
	\and
{	$^1$Chinese Academy of Sciences}
	\and
{	$^2$National University of Singapore}
	\and
{	$^3$Adobe Research}
	\and
{	$^4$Sun Yat-sen University  }
	\and
{    $^5$China-Singapore Institute of Digital Media}
	\and  
	liusi@iie.ac.cn, xdliang328@gmail.com
}


\vspace{-4cm}

\maketitle
	\begin{abstract}
		
		Both parametric and non-parametric approaches have demonstrated encouraging performances in the human parsing task, namely segmenting a human image into several semantic regions (e.g., hat, bag, left arm, face). In this work, we aim to develop  a new solution with the advantages of both methodologies, namely supervision from annotated data and the flexibility to use newly annotated (possibly uncommon) images, and present a quasi-parametric human parsing model. Under the classic K Nearest Neighbor (KNN)-based nonparametric framework, the parametric Matching Convolutional Neural Network (M-CNN) is proposed to predict the matching confidence and displacements of the best matched region in the testing image for a particular semantic region in one KNN image. Given a testing image, we first retrieve its KNN images from the annotated/manually-parsed human image corpus. Then each semantic region in each KNN image is matched with confidence to the testing image using M-CNN, and the matched regions from all KNN images are further fused, followed by a superpixel smoothing procedure to obtain the ultimate human parsing result. The M-CNN differs from the classic CNN~\cite{krizhevsky2012imagenet} in that the tailored cross image matching filters are introduced to characterize the matching between the testing image and the semantic region of a KNN image. The cross image matching filters are defined at different convolutional  layers, each aiming to capture a particular range of displacements. Comprehensive evaluations over a large dataset with 7,700 annotated human images well demonstrate the significant performance gain from the quasi-parametric model over the state-of-the-arts~\cite{yamaguchi2013paper,yamaguchi2012parsing}, for the human parsing task.
		
	\end{abstract}

	\section{Introduction}
	
	\begin{figure}[t]
		\begin{center}
			\includegraphics[width=1\linewidth]{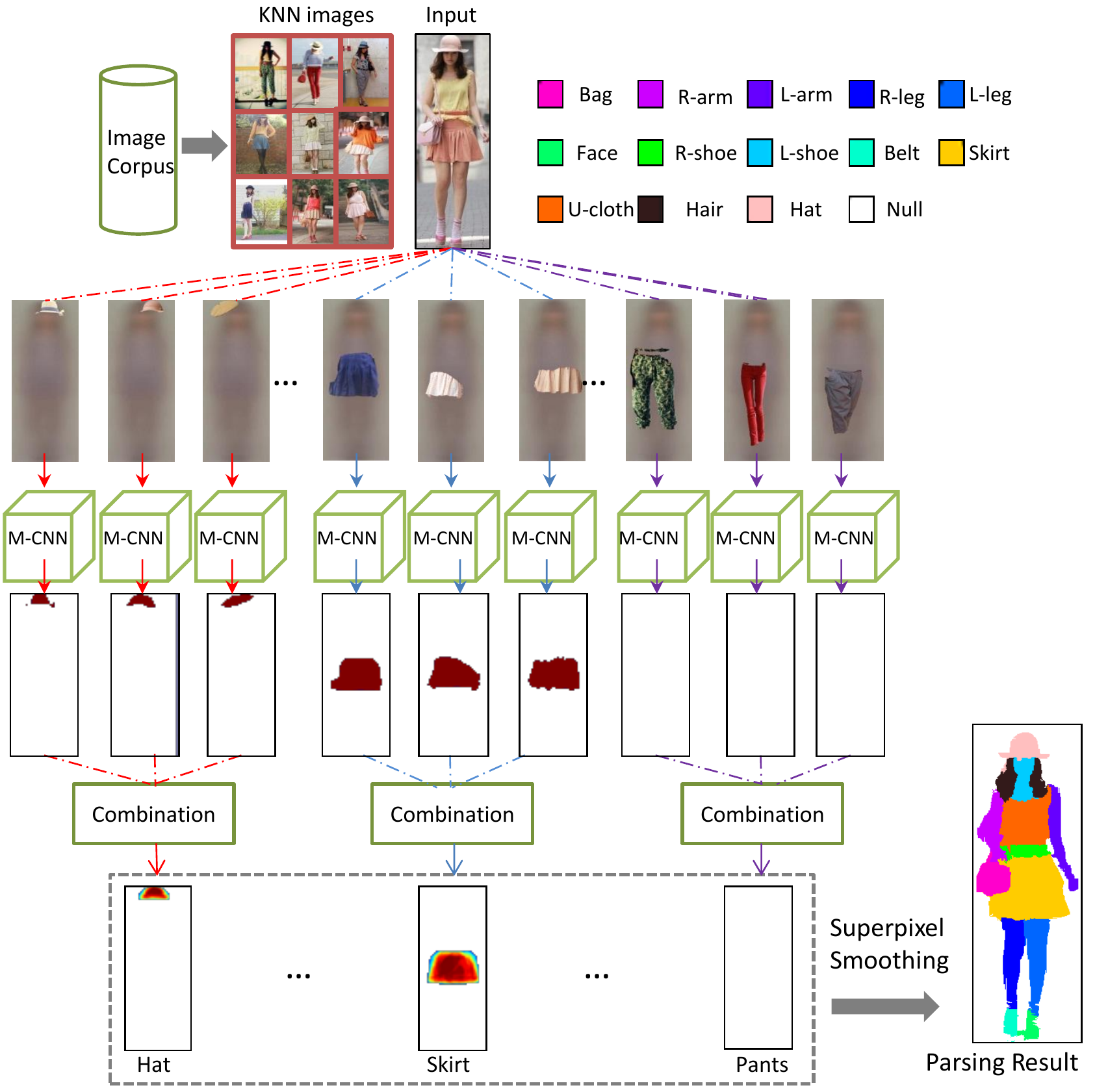}
		\end{center}
		\caption{
			Illustration of our quasi-parametric human parsing framework. 
			Given a testing image,  its K Nearest Neighbour (KNN) images are retrieved from the  manually-annotated image corpus. 
			Then the input image is paired with each semantic region (e.g., hat, skirt and pants) of its KNN images and each pair is fed into the M-CNN individually. The M-CNN predicts the matching confidence and displacements between the input image pair. 
			Then the corresponding label maps are transferred from the KNN region to the testing image. All transferred label maps from different KNN images are combined  to produce a probability map for each label.  The probability map is further refined by superpixel smoothing to generate the  final parsing result. The color legend is shown in the top. For better viewing of all figures in this paper, please refer to the original zoomed-in color pdf file.   
		}
		\label{fig:framework}
		\vspace{-0.2in}
	\end{figure} 
	
	Human parsing, namely partitioning the human body into several semantic regions (e.g., hat, left/right leg, glasses and upper-body clothes), has drawn much attention in  recent years~\cite{yamaguchi2013paper,yamaguchi2012parsing,dong2013deformable,dongtowards} and serves as the basis for many high-level applications, such as clothing classification~\cite{chen2012describing} and retrieval~\cite{liu2012street}.

	\begin{figure*}[t]
		\begin{center}
			\includegraphics[width=0.9\linewidth]{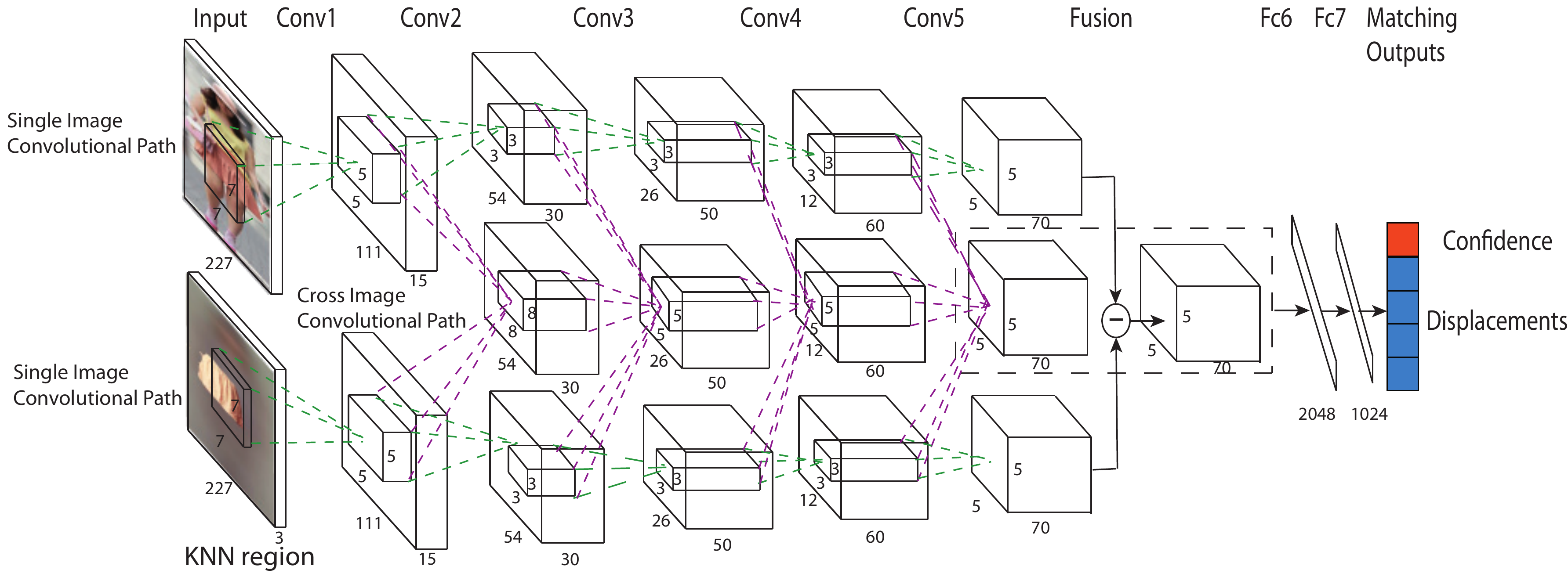}
		\end{center}
		\caption{
			The architecture of the proposed Matching Convolutional Neural Network (M-CNN) with parameters shown. 	
			The input image is cropped from a human image by~\cite{girshick2013rich}, which is paired with the semantic region (e.g., skirt). The image pair is fed into the M-CNN to estimate the 
			matching outputs including $1$-dim matching confidence (red square) and $4$-dim displacements (blue squares). The M-CNN is composed of two kinds of paths.
			In the single image convolutional paths (top and bottom rows), the input and KNN region are independently  convolved  with single image filters (green dashed lines) for hierarchical 		
			feature representation. 
			In the cross image convolutional patch (middle row), the cross image matching filters  (purple dashed lines)  are embedded in Conv2, Conv3, Conv4 and Conv5 layers to achieve multi-ranged matching. Each matching kernel convolves with all features maps of previous convolutional layers. 
			The feature maps of the three paths are  fused to fit the network output. 	}
		\label{fig:cnn}
		\vspace{-0.2in}
	\end{figure*}

	Several parametric and non-parametric  human parsing methods are proposed and show very promising performances on the human parsing task.  On one hand, parametric methods~\cite{yamaguchi2012parsing,dongtowards,liu2014fashion,dong2013deformable}  learn knowledge, such as the appearance of regions of different semantic labels and the structural relationship among different labels, from  annotated images. Those methods usually rely on the manually designed structural models~\cite{dong2013deformable}, which may not fit specific data well and thus achieve only suboptimal performance.  Moreover, for  another set of new training data and semantic labels, new models have to be designed/retrained, which makes those parametric models impractical because new clothing styles may come out quite often.  On the other hand, the non-parametric methods can flexibly use the newly annotated images on the fly and address the issues of parametric model, which are more appealing for practical applications~\cite{liu2011nonparametric}. This kind of methods usually firstly build the pixel-level~\cite{liu2011nonparametric}, superpixel-level~\cite{yamaguchi2013paper}  or hypothesis-level~\cite{tung2014collageparsing,kuettel2012figure,kuettel2012segmentation} matching between a testing image and the annotated images in a corpus, then transfer the labels from the manually annotated images to the testing image  based on the matching outputs, and finally fuse the transferred labels by heuristic aggregation schemes (typically majority voting). However, the quality of matching is usually limited by the lack of explicit semantic meaning of the bottom-up superpixels or hypotheses.

	The above-mentioned parametric and non-parametric human parsing methods rely on the hand designed pipelines composed of multiple sequential components, e.g., hand-crafted feature extraction, bottom-up over-segmentation, human pose estimation, manually designed complex model structure. Therefore, the possibly bad performance of each component may become the bottleneck of 
	the performance of 	the whole pipeline.  For example, the human pose estimation,  an important component in the above pipeline, itself is  quite a challenging task. 
	Such a sequential processing strategy usually makes  the whole pipeline mostly  suboptimal. 
	Instead of a combination of multiple sequential steps, several Convolutional Neural Network (CNN) based 
	methods are proposed for an end-to-end image parsing~\cite{hariharan2014simultaneous,wangdeep,RecurrentICML14,YannLepami13}. 
	However, these deep models  cannot be easily updated to incorporate new semantic labels.

	To address these issues, we propose a quasi-parametric human parsing framework, which inherits the merits of both the parametric and non-parametric models. The proposed end-to-end framework is able to take full advantage of the supervision information from annotation training data, and meanwhile is easy to extend for new added labels. The core part of the proposed framework is a specially designed 
	Matching Convolutional Neural Network (M-CNN) to match any semantic region of a KNN image (also denoted as KNN region in this paper) to the testing image.

	As shown in Fig.~\ref{fig:framework},  we first apply the human detection~\cite{girshick2013rich} to a testing image and obtain the human centric image. Then the K Nearest Neighbours (KNN) images of  the test image  is retrieved from the annotated/manually-parsed image corpus.
	Each KNN image derives several semantic regions, which are generated by masking out the background region with the mean image of the image corpus. Three KNN regions for hat, skirt and pants   are shown 	 in Fig.~\ref{fig:framework}.
	Then, the pair of the test image and   each of the KNN regions is fed into the proposed  M-CNN to 
	estimate their matching confidence and displacements. 
	The   matching confidence measures how the KNN region matches  the input image  while the   displacements describe the coordinate translations between the KNN region and the matched region in the testing image.  	The matching confidences are then averaged over all KNN regions and thresholded  to predict whether a specific  label is present.    For the labels predicted to be present, such as hat and skirt in Fig.~\ref{fig:framework},  the corresponding label maps can be transferred from the 
	KNN regions to the testing image based on the estimated displacements.  For the  labels predicted as invisible, such as pants in Fig.~\ref{fig:framework}, no transferred label map is generated.   
	Then, all matched regions of a specific label are combined to produce a  probability map. Finally, the  probability maps for all labels are refined by a superpixel smoothing step to get the final parsing result.

	Reliable matching between an input image and a KNN region is challenging, because the matching needs to handle the large spatial variance of semantic regions. For example, the bags can be placed on the left, right or in front of the human body.
	The proposed M-CNN is able achieve accurate multi-ranged matching.  As shown in Fig.~\ref{fig:cnn}, M-CNN contains three  paths, i.e., 
	two	single image convolutional paths and  a cross image convolutional path.  The  \emph{single image convolutional path} receives 
	the input image or a particular KNN region, and produces its    discriminative hierarchical  feature representations layer by layer.  The \emph{cross image convolutional path} embeds cross image filters into every convolutional layer to characterize the multi-ranged matching. The cross image filters are applied to all feature maps in previous convolutional layers, including the single image feature maps and cross image feature maps. 
	Because the scale of receptive fields of the feature maps increase when tracing up the M-CNN,   the cross image matching filters capture the displacements  from the near-range to the far-range. Therefore the feature maps from the cross image convolutional path can well represent the displacements. 
	Because the feature maps generated by the two single image convolutional paths  are excellent feature representations, their absolute difference maps  are  calculated as  another measurement of the displacements. The difference maps are combined with the cross image feature maps and then link to the subsequent fully connected layers.  Finally, the matching confidence and displacements are  regressed.    Since the M-CNN targets at matching an input image and any KNN region of any semantic label, it can work even if  new semantic labels are included. 
	Instead of training a M-CNN for each label,  we train a unified M-CNN for  all KNN regions of all labels.

	Comprehensive evaluations over a large dataset with $7,700$ annotated
	human images well demonstrate the effectiveness of our
	quasi-parametric framework. The major contributions are summarized as follows:  
	\begin{itemize}
		\item We build a novel deep quasi-parametric human parsing framework. It can learn from  annotated data
		and   also flexibly use newly annotated (possibly uncommon) 	images.
		\item 	We propose a Matching Convolutional Neural Network (M-CNN) to 
		match  a  semantic region of a KNN image to a testing image. The novel cross image filters are embedded into different convolutional  layers, each aiming to capture a particular range of displacements.
		\item We integrate all the step-by-step components (over-segmentation, pose estimation, feature extraction, label modeling, etc.) in traditional pipelines into one unified end-to-end deep CNN framework.
	\end{itemize}

	\section{Related Work}
	In this section, we sequentially review the parametric human parsing methods, non-parametric methods and deep learning based  methods. 
	
	For parametric human parsing, Yamaguchi et al.~\cite{yamaguchi2012parsing} proposed to boost the human parsing with pixel-level classification by relying on human pose estimation. To capture more complex contextual information, Dong et al.~\cite{dong2013deformable} designed an And-or Graph structure to model the correlations of a group of parselets, and their extension work~\cite{dongtowards}  unified  the human parsing and pose estimation  in one framework. The image co-segmentation and region co-labeling for human parsing were also used to capture the correlations between different human images~\cite{yangclothing}.
	In addition, Liu et al.~\cite{liu2014fashion} utilized user-generated category tags to build a human parser. 
	For general image parsing, Tighe et al.~\cite{tighe2013finding} proposed a segmentation by detection approach. Firstly, the bounding boxes of the objects are estimated by exemplar SVM~\cite{tighe2013finding}, based on which the segmentation masks are transferred from  the image corpus to the input image. 	In general, the power of existing parametric methods is largely limited by the suboptimal performance of many hand designed intermediate components, such as pose estimation, and also cannot be easily  extended to parse new labels.

	In non-parametric human parsing, pixels~\cite{liu2011nonparametric}, superpixels~\cite{tighe2010superparsing,gould2014superpixel,kuettel2012segmentation} and  object proposals~\cite{tung2014collageparsing,kuettel2012figure,kuettel2012segmentation,serrano2013predicting} were used to facilitate non-parametric image parsing. Specifically, the model of Yamaguchi et al.~\cite{yamaguchi2013paper}  transferred parsing masks from retrieved examples to the query image. Their label transferring is based on superpixels, which are generated by over-segmentation  based on  the low level appearance cues and therefore lack
	semantic meaning. Liu et al.~\cite{liu2011nonparametric}  used SIFT Flows to build the pixel-pixel correspondence and the dense deformation field between images. However, the optimization problem for finding the SIFT flow is rather complex and expensive to solve.  
	Recently, Long et al.~\cite{zbontar2014computing}  proved  the better performance of convolutional  activation features over traditional features, such as SIFT,  for tasks requiring correspondence. Overall, the non-parametric methods are limited by the inaccurate matching, which results in  the noises/outliers during the label transferring.

	Our quasi-parametric model integrates the advantages of both parametric models and non-parametric models by  the proposed M-CNN. There  exist some works on semantic segmentation with CNN architectures. Girshick et al.~\cite{girshick2013rich} and its extension work~\cite{hariharan2014simultaneous} proposed to classify the candidate regions by CNN for semantic segmentation. Wang et al.~\cite{wangdeep} presented a joint task learning framework, in which the object localization task and the object segmentation task are tackled collaboratively  via CNN.  Farabet et al.~\cite{YannLepami13} trained a multi-scale CNN from raw pixels to extract deep features for assigning the label to each pixel. The recurrent CNN~\cite{RecurrentICML14} was proposed to speed up scene parsing and achieved the state-of-the-art performance. 
	Our M-CNN inherits the merit of existing CNN parsing models in our single image convolutional path. 
	It differs from all existing CNN based parsing models in that we handle a pair of images instead of a single image and we incorporate cross image filters to specifically characterize multi-ranged matching. Last but not least,  M-CNN can effortlessly handle  new semantic labels.

	\section{ Quasi-parametric Human Parsing }

	For each human image,  we first retrieve its KNN images from the annotated image corpus (Sec.~\ref{sec:knn}). Then, M-CNN predicts the matching confidence and displacements between the input image and a semantic region from one KNN image, based on which a label map is generated (Sec.~\ref{sec:M-CNN}).   Finally, all label maps are fed into a post processing procedure to  produce the  parsing result (Sec.~\ref{sec:refine}). 
	
	\subsection{K-Nearest-Neighbor Retrieval}  \label{sec:knn}
	For each of the input images, we  use the human detection algorithm~\cite{girshick2013rich} to detect the human body.
	The resulting  human centric image $I$ is then rescaled  to  $227\times227\times3$.
	We then extract a global $4,096$-dimensional feature  
	from the penultimate fully-connected layer in the pre-trained CNN model trained on ILSVRC 2012 classification dataset based on the Krizhevsky architecture  \cite{krizhevsky2012imagenet}. Its KNN images $G = \left\{ {{g_1},{g_2},...,{g_K}} \right\}$ are retrieved from the image corpus based on the deep features.

	\subsection{ Matching Convolutional Neural Network }  \label{sec:M-CNN}
	
	\textbf{Input, Output and Loss Function:}
	Given a  label $l \in \{ 1,...,L\}$, where $L$ is the total number of  labels, the input image $I$ and each KNN regions $g_{kl}$ from KNN image $g_k$  form a pair and are fed into the M-CNN. 
	Note that $g_k$ is from the image corpus and thus its label map is known. 
	$g_{kl}$ is generated by  keeping the regions of the label $l$ in $g_k$ and  masking out other regions by the mean image calculated from the image corpus.
	If $g_k$ does not contain  label $l$, $g_{kl}$ is exactly the mean image.

	Given an image pair $\{{I},g_{kl}\}$, M-CNN learns a  regressor   to estimate the matching confidence and displacements between them. 
	The  $1$-dim {matching confidence} $c_{kl}$ indicates how well  $g_{kl}$ can be matched to $I$.  In the trainig phase, it is a binary index indicating whether the  KNN region is matched to the input image. 
	$g_{kl}$  is considered to be matched with $I$ if and only if $I$ contains the label $l$. 
	In the testing phase, higher value of $c_{kl}$ indicates that better matching is found. 
	We denote the coordinates of the upper  left and lower right corner of the KNN region in  $g_{kl}$ as $U_g=[{g_{kl}^{x_1}},{g_{kl}^{y_1}}]$ and $W_g=[{g_{kl}^{x_2}},{g_{kl}^{y_2}}]$.  Similarly, the coordinates of the matched region  in  $I$ are denoted as $U_I=[{I^{x_1}},{I^{y_1}}]$ and $W_I=[{I^{x_2}},{I^{y_2}}]$.  The coordinates are normalized by the height and width of the image into the range  $[0,1]$. 
	The $4$-dim  displacements ${{t}_{kl}}$ represents the differences between 
	$U_g$ and $U_I$, $W_g$ and $W_I$. 
	Instead of using a classification loss~\cite{krizhevsky2012imagenet}, we train the M-CNN  by minimizing
	${\ell _2}$ distance between the ground truth   $[{{{c}}_{kl}},{{t}}_{kl}]$
	and the prediction $[{{\tilde{c}}_{kl}},{\tilde{{t}}}_{kl}]$.
	The corresponding $\ell_2$ loss for each image $I$ is defined as
	\begin{equation}
		\begin{array}{l}
			J = \frac{1}{{L \times K}}\sum\limits_{k = 1}^K {\sum\limits_{l = 1}^L {{{\left\| {{{ c}_{kl}} - {{\tilde c}_{kl}}} \right\|}^2}} } \\
			+ \frac{1}{{L \times K}}\sum\limits_{k = 1}^K {\sum\limits_{l \in \varphi \left( {{g_k}} \right) \cap \varphi \left( I \right) \cap \{ 1,...,L\} } {{{\left\| {{{ t}_{kl}} - {{\tilde t}_{kl}}} \right\|}^2}} } )
		\end{array}
		\label{eq:loss}
	\end{equation}
	where 	$\varphi \left(  \cdot  \right)$  is the label set contained in the specific image. 
	The first term in Eq.~\eqref{eq:loss} is the loss for matching confidence while the second term corresponds to the loss for displacements. 
	We  penalize the displacements loss when both  the KNN image $g_k$ and the input image $I$ contain the label $l$.
	Then the losses of all training pairs  are summed and the parameters are learned by  back propagation.

	\begin{figure}[h]
		\begin{center}
			\includegraphics[width=1\linewidth]{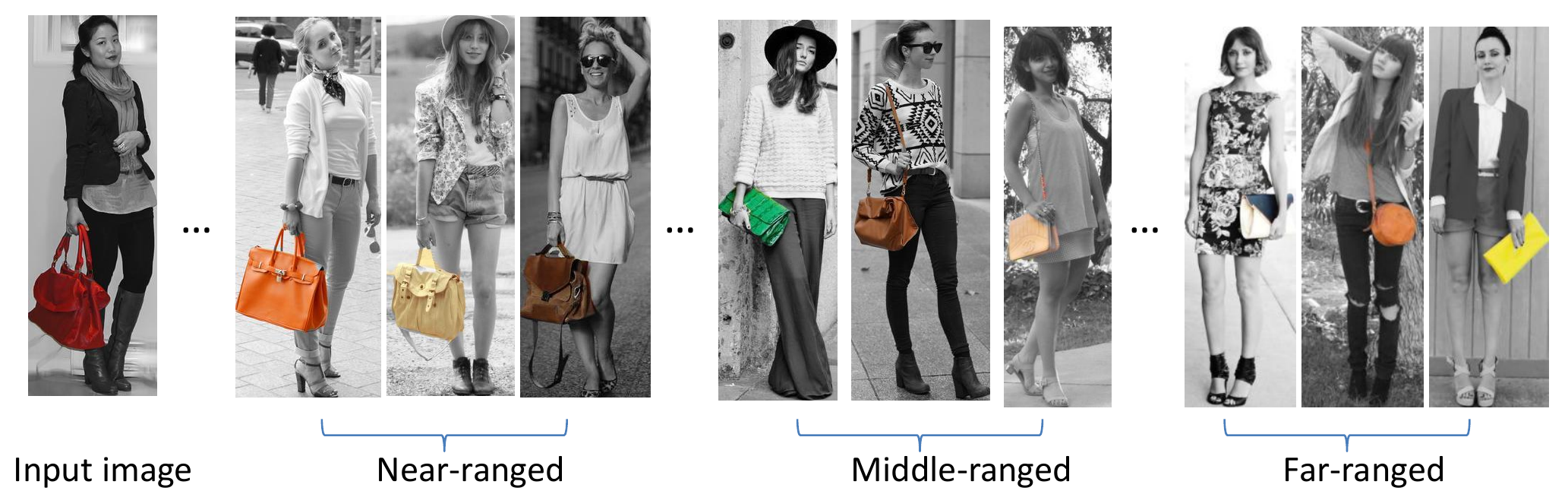}
		\end{center}
		\vspace{-0.1in}
		\caption{ Illustration of the multi-range matching. For a specific label, e.g., bag, the  matchings between the input image and KNN regions  may be near/middle/far-ranged. To facilitate the display, we only keep the bags regions unchanged and gray other pixels.
		}
		\label{fig:omini_range_matching}
		\vspace{-0.1in}
	\end{figure} 
	
	\textbf{Architecture:}
	Since the KNN images are retrieved based on the global appearance similarity, the KNN region of each label may locate quite differently in images. For example, in Fig.~\ref{fig:omini_range_matching}, the bags can be placed on the left side or right side or in front of the human body.   M-CNN is designed to  estimate the multi-ranged matching by embedding the  cross image matching filters in different convolutional layers.
	As shown in Fig.~\ref{fig:cnn}, M-CNN contains two kinds of  paths, i.e., two single
	image convolutional paths and one cross image convolutional path the outputs of these three paths are further fused to estimate the matching confidence and displacements.
	The single image path aims for hierarchical  feature representation 	while the cross image convolutional path estimates the displacements between the input pair.

	\emph{Single Image Convolutional Path:} We have two instantiations of the single image path in the top and bottom row of Fig.~\ref{fig:cnn}, each of which separately   processes  $I$ or  $g_{kl}$. 
	They share the same architecture and extract the hierchical feature representations of $I$ or $g_{kl}$. 
	The outputs are their respective feature maps in ``conv5''.
	In this path, the single image filters of the next convolutional layers are connected to those feature maps in the previous layer, shown as the green dashed line in Fig.~\ref{fig:cnn}. The ReLU non-linearity is applied to the output of every convolutional layer. 
	The sizes of feature maps  are gradually reduced 	by using the stride of $2$ for all the convolutional  layers.
	The most important difference between M-CNN and the infrastructure in~\cite{krizhevsky2012imagenet} is that M-CNN removes the pooling layer. Although pooling is useful for enhancing translation invariance for object recognition, it loses precise spatial information that is necessary for accurately predicting the locations of the labels~\cite{ jain2013learning}.  
	The details about the network parameters, such as image/feature map sizes, kernel size/numbers are shown in  Fig.~\ref{fig:cnn}. 
	The powerful representation capability of the single image convolutional path lays the foundation for the accurate estimation of the matching confidence and displacements.

	%

	\emph{ Cross Image Convolutional Path:} 
	The cross image convolutional path lies in the middle  row of Fig.~\ref{fig:cnn}. 
	It  outputs the cross image feature maps in ``conv5'' layer. 
	The $m$-th cross image feature map in the $j$-th layer ${x_{j,m}^C}$ is generated by convolving the corresponding matching  filter (including three components ${f_{j,p,m}^I}$, ${f_{j,q,m}^R}$ and ${f_{j,t,m}^C}$) 	with both 	singe image (${x_{j-1,p}^{{I}}}$,  ${x_{j - 1,q}^{{R}}}$) and cross image  feature maps (${x_{j - 1,t}^{{C}}}$) in the $j-1$-th layer. 
	The ${f_{j,p,m}^I}$ component links  the  $p$-th (out of all $P$) input image feature map ${x_{j-1,p}^{{I}}}$ in the $j-1$-th layer to ${x_{j,m}^C}$.
	Analogously, the  ${f_{j,q,m}^R}$ component links the $q$-th (out of all $Q$) KNN region feature map ${x_{j - 1,q}^{{R}}}$ to ${x_{j,m}^C}$.
	Moreover, the ${f_{j,t,m}^C}$ component links the $t$-th (out of all $T$) cross image feature map ${x_{j - 1,t}^{{C}}}$ to ${x_{j,m}^C}$.
	The operation of the matching filters  from one layer to the next is shown as the purple dashed line in Fig.~\ref{fig:cnn}. 
	Mathematically, the cross feature map $ {x_{j,m}^C}$  is calculated by: 
	\begin{equation}
		\begin{array}{l}
			\begin{array}{l}
				{x_{j,m}^C} = \max (0,{b_{j,m}} + \sum\limits_{p = 1}^P {f_{j,p,m}^I*x_{j - 1,p}^I} \\
				\;\;\;\;\;\;\;\;\;\;\;\;\;\;\;\;\;\;\;  + \sum\limits_{q = 1}^Q {f_{j,q,m}^R*x_{j - 1,q}^R}  + \sum\limits_{t = 1}^T {f_{j,t,m}^C*x_{j - 1,t}^C} )
			\end{array}
		\end{array}	\label{eq:cross_img_kernel}
	\end{equation}
	where  $*$ denotes convolution and ${b_{j,m}}$  is the bias for the $m$-th output map. $\max (0, \cdot )$ is the non-linear activation function, and is operated element-wisely. From Eq.~\eqref{eq:cross_img_kernel}, we can see that the cross image feature map in the next layer is calculated by considering both single and cross image feature maps in the previous layer, and thus the displacements between  the input image $I$ and  KNN region $g_{kl}$ can be effectively estimated. 
	Note that along with the M-CNN, the receptive fields of different layers of the single image and cross image convolutional path gradually increase~\cite{visualization13}. 
	In this way, multi-ranged matchings can be achieved.

	Finally, we fuse the feature maps from two single image paths and one cross image path. 
	More specifically, the absolute differences of the feature maps of the input image and the KNN  region (from the single image convolutional path) are first calculated and then  are stacked with the output of the cross image convolutional path. 
	Our fusion is applied on the feature maps. It  is different from  ``Siamese''~\cite{chopra2005learning} architecture which calculates the absolute differences of  the  fully-connected representations. Experiments show  the  our fusion strategy outperforms ``Siamese'' by keeping more spatial information.

	\vspace{-0.1in}
	
	\subsection{Post Processing} \label{sec:refine}
	
	Given the  matching confidences and displacements estimated by M-CNN,  the parsing result of the  input image can be calculated  as follows. 
	Firstly,  the confidence of  $I$ containing the $l$-th label  is calculated by
	averaging matching confidence	${c_{kl}}$ for all KNN regions satisfying  $l \in \varphi \left( {{g_k}} \right)$. 
	If the confidence is greater than a threshold ${\xi _1}$, the label $l$ is predicted as visible in the input image, otherwise predicted as invisible.
	Secondly, we estimate the locations of the visible labels.  
	More specifically,    the coordinates of the 
	matching region  in the input image $I$ is calculated based on  the
	matching displacements ${t_{kl}}$ and the ground-truth coordinates of $g_{kl}$. Then, 
	we morph the associated ground-truth label mask of $g_{kl}$ into  matched region in $I$.
	In this way, we get a probability map $M_l$ of $I$ for each label $l \in [1,L]$. 
	We pixel-wisely max all  $M_l$  for all the labels and get the foreground probability.  The pixels with the probability larger than a threshold ${\xi _2}$ are regarded as the rough foreground, while the remaining are the rough background.  The rough foreground and background are further eroded by a filter size $10$ to produce the final foreground and background seeds. Based on the seeds, we can obtain the background probability by the algorithm~\cite{gulshan2010geodesic}. The obtained background probability  is combined with the foreground probability map  $M_l$, $l \in [1,L]$, based on which, we can get an initial human parsing results with the pixel-wise Maximum a Posterior Probability (MAP) assignment.
	Finally, to respect boundaries of actual semantic labels, we further over-segment  $I$ using the entropy rate based segmentation algorithm~\cite{liu2011entropy} 
	and  assign the label of the superpixel by the majority of its covered pixels' initial parsing results.

	\vspace{-0.1in}
	\section{Experiments}

	\subsection{Experimental Settings}

	\textbf{Datasets}: 	We use the dataset in \cite{liang2015deep} pixel-wisely labeled by  the  $18$ categories defined by Daily Photos dataset~\cite{dong2013deformable}. 
	The dataset contains $7,700$ images ($6,000$ for training, $700$ for  validation and $1,000$ for testing). We adopt four evaluation metrics, i.e., accuracy, average precision, average recall, and average F-1 scores over pixels~\cite{yamaguchi2012parsing}.

	\begin{table*}[htbp]\setlength{\tabcolsep}{1pt}
		\centering
		\caption{Comparison of parsing performances with several architectural variants of our model (cross image matching filters embedded into different convolutional layers, with and without superpixel smoothing) and two state-of-the-arts. }
		\label{experiment_all_label}
		\small{
			\begin{tabular}{cccccccccccccccccccccc}
				\toprule
				\textbf{Method} &\hspace{+0.1in}   \textbf{Accuracy}   &\hspace{+0.1in}   \textbf{F.g. accuracy}  &\hspace{+0.1in}   \textbf{Avg. precision}   &\hspace{+0.1in}     \textbf{Avg. recall}  &\hspace{+0.1in}   \textbf{Avg. $F_1$ score} \\
				\midrule
				Yamaguchi et al.~\cite{yamaguchi2012parsing}  &\hspace{+0.1in}  84.38 &\hspace{+0.1in}  55.59 &\hspace{+0.1in}  37.54 &\hspace{+0.1in}  51.05 &\hspace{+0.1in}  41.80 \\
				PaperDoll~\cite{yamaguchi2013paper}  &\hspace{+0.1in}  88.96 &\hspace{+0.1in}  62.18 &\hspace{+0.1in}  52.75 &\hspace{+0.1in}  49.43 &\hspace{+0.1in}  44.76 \\
				\midrule
				{Siamese}~\cite{chopra2005learning} &\hspace{+0.1in}  {85.24} &\hspace{+0.1in}  {56.42} &\hspace{+0.1in}  {50.27}
				&\hspace{+0.1in}  {48.88} &\hspace{+0.1in}  {47.08}\\
				{M-CNN (w/o cross) } &\hspace{+0.1in}  {88.30} &\hspace{+0.1in}  {69.84} &\hspace{+0.1in}  {58.63}
				&\hspace{+0.1in}  {59.52} &\hspace{+0.1in}  {56.99}\\
				{M-CNN  (cross 5 )} &\hspace{+0.1in}  {88.62} &\hspace{+0.1in}  {69.88} &\hspace{+0.1in}  {60.89}
				&\hspace{+0.1in}  {60.47} &\hspace{+0.1in}  {58.07}\\
				{M-CNN (cross 5,4 )} &\hspace{+0.1in}  {89.41} &\hspace{+0.1in}  {72.44} &\hspace{+0.1in}  {58.93}
				&\hspace{+0.1in}  {63.16} &\hspace{+0.1in}  {60.03}\\
				{M-CNN (cross 5,4,3 )} &\hspace{+0.1in}  {88.97} &\hspace{+0.1in}  {70.84} &\hspace{+0.1in}  {60.27}
				&\hspace{+0.1in}  {62.23} &\hspace{+0.1in}  {60.36}\\
				{M-CNN} &\hspace{+0.1in}  \textbf{89.57} &\hspace{+0.1in}  \textbf{73.98} &\hspace{+0.1in}  \textbf{64.56}
				&\hspace{+0.1in}  {65.17} &\hspace{+0.1in}  \textbf{62.81}\\
				{M-CNN (cross 5,4,3,2,1 )} &\hspace{+0.1in}  {89.42} &\hspace{+0.1in}  {71.86} &\hspace{+0.1in}  {63.13}
				&\hspace{+0.1in}  {63.49} &\hspace{+0.1in}  {61.53}\\
				{M-CNN(w/0 ss )} &\hspace{+0.1in}  {87.08} &\hspace{+0.1in}  {71.73} &\hspace{+0.1in}  {55.88}
				&\hspace{+0.1in}  \textbf{65.32} &\hspace{+0.1in}  {59.39}\\								       	  	 
				\bottomrule
			\end{tabular}%
		}
	\end{table*}

	\begin{table*} [htbp]\setlength{\tabcolsep}{2pt}
		\small
		{
			\centering
			{
				\caption{$F_1$ scores of foreground semantic labels. Comparison of $F_1$-scores with several architectural variants of our model and two state-of-the-art methods.}
				\label{experiment_each_label}
				\begin{tabular}{cccccccccccccccccccccc}
					\toprule
					Method &\hspace{-0.06in}Hat&\hspace{-0.06in} 	Hair&\hspace{-0.06in} 	S-gls&\hspace{-0.06in} 	U-cloth&\hspace{-0.06in} 	Skirt&\hspace{-0.06in} 	Pants&\hspace{-0.06in} 	Dress&\hspace{-0.06in} 	Belt&\hspace{-0.06in} 	L-shoe&\hspace{-0.06in} 	R-shoe&\hspace{-0.06in} 	Face&\hspace{-0.06in} 	L-leg	&\hspace{-0.06in}  R-leg&\hspace{-0.06in} 	L-arm&\hspace{-0.06in} 	R-arm &\hspace{-0.06in}  Bag	&\hspace{-0.06in}  Scarf\\
					\midrule
					Yamaguchi et al.~\cite{yamaguchi2012parsing} &\hspace{-0.06in} 8.44&\hspace{-0.06in} 59.96&\hspace{-0.06in} 12.09&\hspace{-0.06in} 56.07&\hspace{-0.06in} 17.57&\hspace{-0.06in} 55.42&\hspace{-0.06in} 40.94&\hspace{-0.06in} 14.68&\hspace{-0.06in} 38.24&\hspace{-0.06in} 38.33&\hspace{-0.06in} 72.1&\hspace{-0.06in} 58.52&\hspace{-0.06in} 57.03&\hspace{-0.06in} 45.33&\hspace{-0.06in} 46.65&\hspace{-0.06in} 24.53&\hspace{-0.06in} 11.43\\			
					PaperDoll~\cite{yamaguchi2013paper}  &\hspace{-0.06in}1.72&\hspace{-0.06in} 63.58&\hspace{-0.06in} 0.23&\hspace{-0.06in} 71.87&\hspace{-0.06in} 40.2&\hspace{-0.06in} 69.35&\hspace{-0.06in} 59.49&\hspace{-0.06in} 16.94&\hspace{-0.06in} 45.79&\hspace{-0.06in} 44.47&\hspace{-0.06in} 61.63&\hspace{-0.06in} 52.19&\hspace{-0.06in} 55.6&\hspace{-0.06in} 45.23&\hspace{-0.06in} 46.75&\hspace{-0.06in} 30.52&\hspace{-0.06in} 2.95 \\
					\midrule
					{Siamese}~\cite{chopra2005learning}  &\hspace{-0.06in}74.02&\hspace{-0.06in} 	44.70&\hspace{-0.06in} 	11.78&\hspace{-0.06in} 	65.69&\hspace{-0.06in} 	74.69&\hspace{-0.06in} 	61.52&\hspace{-0.06in} 	70.30&\hspace{-0.06in} 	1.85&\hspace{-0.06in} 	37.12&\hspace{-0.06in} 	34.16&\hspace{-0.06in} 	39.36&\hspace{-0.06in} 	50.24&\hspace{-0.06in} 	51.65&\hspace{-0.06in} 	32.54&\hspace{-0.06in} 	30.59&\hspace{-0.06in} 	24.23&\hspace{-0.06in} 	39.37\\
					{M-CNN (w/o cross) } &\hspace{-0.06in}77.67&\hspace{-0.06in} 	63.11&\hspace{-0.06in} 	29.68&\hspace{-0.06in} 	72.26&\hspace{-0.06in} 	73.66&\hspace{-0.06in} 	\textbf{70.92}&\hspace{-0.06in} 	78.79&\hspace{-0.06in} 	25.51&\hspace{-0.06in} 	43.25&\hspace{-0.06in} 	44.53&\hspace{-0.06in} 	67.99&\hspace{-0.06in} 	60.90&\hspace{-0.06in} 	64.90&\hspace{-0.06in} 	47.35&\hspace{-0.06in} 	39.78&\hspace{-0.06in} 	41.14&\hspace{-0.06in} 	40.06\\
					{M-CNN (cross 5) } &\hspace{-0.06in}67.61&\hspace{-0.06in} 	65.30&\hspace{-0.06in} 	38.51&\hspace{-0.06in} 	69.06&\hspace{-0.06in} 	73.95&\hspace{-0.06in} 	69.33&\hspace{-0.06in} 	81.88&\hspace{-0.06in} 	28.41&\hspace{-0.06in} 	46.78&\hspace{-0.06in} 	41.01&\hspace{-0.06in} 	70.82&\hspace{-0.06in} 	59.49&\hspace{-0.06in} 	66.28&\hspace{-0.06in} 	52.61&\hspace{-0.06in} 	39.73&\hspace{-0.06in} 	40.89&\hspace{-0.06in} 	46.01\\
					{M-CNN (cross 5,4) } &\hspace{-0.06in}76.42&\hspace{-0.06in} 	65.91&\hspace{-0.06in} 	{46.97}&\hspace{-0.06in} 	71.51&\hspace{-0.06in} 	74.29&\hspace{-0.06in} 	68.43&\hspace{-0.06in} 	82.16&\hspace{-0.06in} 	41.31&\hspace{-0.06in} 	41.83&\hspace{-0.06in} 	43.03&\hspace{-0.06in} 	73.27&\hspace{-0.06in} 	59.35&\hspace{-0.06in} 	62.18&\hspace{-0.06in} 	52.63&\hspace{-0.06in} 	\textbf{54.04} &\hspace{-0.06in} 	46.25&\hspace{-0.06in} 	40.58\\
					{M-CNN (cross 5,4,3) } &\hspace{-0.06in} 79.38&\hspace{-0.06in} 	\textbf{67.64}&\hspace{-0.06in} 	34.50&\hspace{-0.06in} 	70.72&\hspace{-0.06in} 	76.57&\hspace{-0.06in} 	69.17&\hspace{-0.06in} 	\textbf{83.81}&\hspace{-0.06in} 	25.69&\hspace{-0.06in} 	\textbf{54.82}&\hspace{-0.06in} 	44.61&\hspace{-0.06in} 	\textbf{75.22}&\hspace{-0.06in} 	\textbf{63.44}&\hspace{-0.06in} 	67.74&\hspace{-0.06in} 	54.94&\hspace{-0.06in} 	48.59&\hspace{-0.06in} 	40.10&\hspace{-0.06in} 	46.40\\
					{M-CNN}  &\hspace{-0.06in}\textbf{80.77} &\hspace{-0.06in} 	65.31&\hspace{-0.06in} 	35.55&\hspace{-0.06in} 	\textbf{72.58}&\hspace{-0.06in} 	77.86&\hspace{-0.06in} 	70.71&\hspace{-0.06in} 	81.44&\hspace{-0.06in} 	38.45&\hspace{-0.06in} 	53.87&\hspace{-0.06in} 	\textbf{48.57}&\hspace{-0.06in} 	72.78&\hspace{-0.06in} 	63.25&\hspace{-0.06in} 	\textbf{68.24}&\hspace{-0.06in} 	\textbf{57.40}&\hspace{-0.06in} 	{51.12}&\hspace{-0.06in} 	\textbf{57.87}&\hspace{-0.06in} 	43.38\\
					{M-CNN (cross 5,4,3,2,1) } &\hspace{-0.06in}77.34&\hspace{-0.06in} 	65.56&\hspace{-0.06in} 	45.19&\hspace{-0.06in} 	70.64&\hspace{-0.06in} 	\textbf{78.64}&\hspace{-0.06in} 	69.95&\hspace{-0.06in} 	82.72&\hspace{-0.06in} 	\textbf{46.52}&\hspace{-0.06in} 	45.72&\hspace{-0.06in} 	44.45&\hspace{-0.06in} 	71.27&\hspace{-0.06in} 	61.59&\hspace{-0.06in} 	63.49&\hspace{-0.06in} 	53.23&\hspace{-0.06in} 	50.72&\hspace{-0.06in} 	55.96&\hspace{-0.06in} 	45.48 \\
					{M-CNN(w/o ss) } &\hspace{-0.06in}72.26&\hspace{-0.06in} 	61.32&\hspace{-0.06in} 	\textbf{49.89}&\hspace{-0.06in} 	68.74&\hspace{-0.06in} 	74.80&\hspace{-0.06in} 	66.41&\hspace{-0.06in} 	78.24&\hspace{-0.06in} 	40.13&\hspace{-0.06in} 	47.86&\hspace{-0.06in} 	43.86&\hspace{-0.06in} 	67.66&\hspace{-0.06in} 	51.61&\hspace{-0.06in} 	59.37&\hspace{-0.06in} 	50.34&\hspace{-0.06in} 	44.38&\hspace{-0.06in} 	50.78&\hspace{-0.06in} 	\textbf{46.04}\\
					\bottomrule
				\end{tabular}
			}
		}
		\vspace{-0.2in}
	\end{table*}

	\textbf{Training Image Pairs Generation}:  To reduce over-fitting in the model training and partially address the detection error,  we enlarge the cropped human centric images  region  by  $1$ and $1.2$ times.
	We also horizontally  mirror the images. In short, each image has $4$ variations and training data can be greatly augmented.

	For each of $6,000$ training images, $50$ KNN images are retrieved from the image corpus.  
	Each training image and each KNN region  form a training pair. 
	After unevenly sampling of the training pairs to   balance  different labels, we finally  have  $5$ million pairs, which even outnumbers the 
	that of ILSVRC2012~\cite{krizhevsky2012imagenet}. We  shuffle the training pair in order to increase the diversity of each epoch.

	\textbf{Implementation Details}: We implement the M-CNN under the Caffe framework~\cite{jia2014caffe} and train it using stochastic gradient descent with a batch size of $128$ examples, momentum of $0.9$, and
	weight decay of $0.0005$. We use an equal learning rate for all layers. 
	The learning rate is adjusted  manually  by dividing $10$ when the validation error rate stops decreasing with the current learning rate. The learning rate is initialized at $0.0005$. We train M-CNN for roughly $50$ epochs, which takes $11$ to $12$ days on one NVIDIA GTX TITAN 6GB GPU.
	In the training phase, we first calculate the element-wise mean and variance of the matching confidence and displacements 	  of the whole image corpus, 
	and  element-wisely normalize the training output by  the  mean and variance.
	In the testing phase, we   project   the matching confidence and displacements estimated by M-CNN to their absolute values  by the mean and variance. 		 
	In the post processing step, the thresholds ${\xi _1}$ and ${\xi _2}$ are set to be $0.8$ and $0.5$.
	The number of KNN regions for each input image is set as $9$.

	%
	
	\subsection{Results and Analysis}

	\textbf{Comparison with The State-of-the-arts}: We compare  our  M-CNN based quasi-parametric human parsing framework with  two state-of-the-arts: Yamaguchi et al.~\cite{yamaguchi2012parsing}   and PaperDoll~\cite{yamaguchi2013paper}. We use their publicly available codes and train their models with the same $6,000$ training images as our method for  fair comparison. We do not compare with  Dong et al.~\cite{dong2013deformable} because their code is not publicly available and their method is reported to be slower than ours.

	The average results for all labels are in Table \ref{experiment_all_label}.
	The methods of Yamaguchi et al.~\cite{yamaguchi2012parsing} and the PaperDoll~\cite{yamaguchi2013paper} trained on the same $6,000$ training images and tested on the same $1,000$ images as M-CNN, and their average $F_1$ scores achieve  $41.80\%$ and $44.76\%$. 
	Our ``M-CNN"  significantly outperforms these two baselines by over $21.01\%$ for Yamaguchi et al.~\cite{yamaguchi2012parsing} and 
	$18.05\%$  for PaperDoll~\cite{yamaguchi2013paper}.
	``M-CNN'' also gives a huge boost in foreground accuracy: the two baselines achieve $55.59\%$ for Yamaguchi et al.~\cite{yamaguchi2012parsing} and $62.18\%$ for PaperDoll~\cite{yamaguchi2013paper} while ``M-CNN" obtains $73.98\%$. 
	``M-CNN" also obtains much higher precision (64.56\% vs 37.54\% for~\cite{yamaguchi2012parsing} and 52.75\% for~\cite{yamaguchi2013paper}) as well as higher recall (65.17\% vs 51.05\% for~\cite{yamaguchi2012parsing} and 49.43\% for~\cite{yamaguchi2013paper}). 
	This verifies the effectiveness of our end-to-end M-CNN based quasi-parametric framework.

	\begin{figure*}[t]
		\begin{center}
			\includegraphics[width=0.9\linewidth]{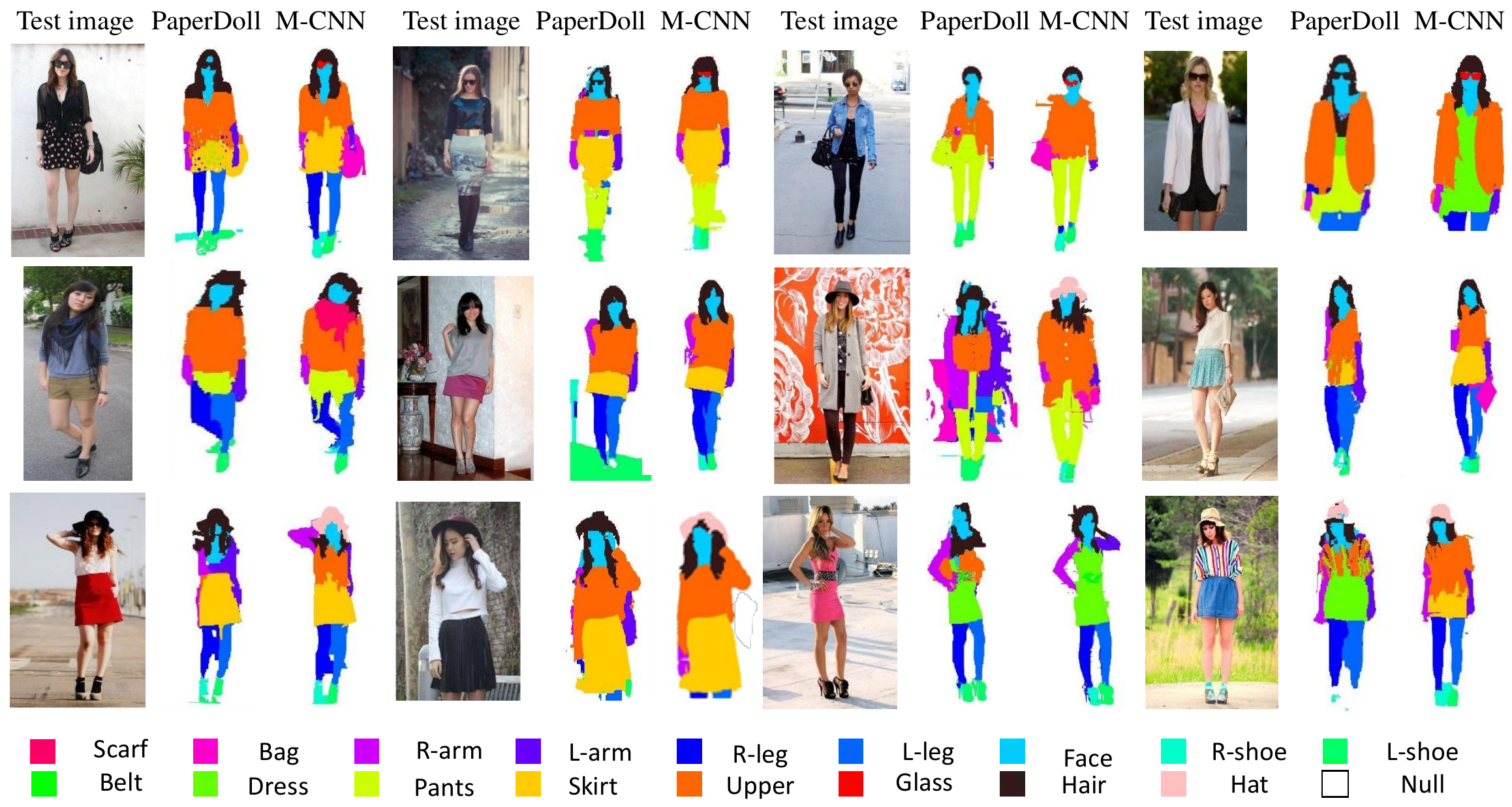}
		\end{center}
		\vspace{-0.1in}
		\caption{Comparison of our parsing results with the PaperDoll Method. For each image,
			we show the testing image, parsing results by PaperDoll~\cite{yamaguchi2013paper}, our ``M-CNN'' sequentially.	}
		\label{fig:final_fig}
		\vspace{-0.2in}
	\end{figure*} 
	
	We also present the F1-scores for each label in Table \ref{experiment_each_label}. 
	Generally, ``M-CNN''  shows much higher performance than the baselines.
	In terms of predicting labels for small semantic  regions such as hat, belt, bag and scarf, our method achieves a large gain, e.g. 
	43.38\% vs 11.43\%~\cite{yamaguchi2012parsing} and 2.95\%~\cite{yamaguchi2013paper} for scarf, 
	57.87\% vs 24.53\%~\cite{yamaguchi2012parsing}, 30.52\%~\cite{yamaguchi2013paper} for bag
	and 
	38.45\% vs 14.68\%~\cite{yamaguchi2012parsing} and 16.94\%~\cite{yamaguchi2013paper} for belt. 
	It demonstrates that our quasi-parametric network can effectively capture the internal relations between the labels and robustly predict the label masks with various clothing styles and poses.

	\textbf{Ablation of Our Networks}:
	We also extensively explore different CNN architectures 
	to demonstrate the effectiveness of each component in M-CNN more transparently.  The architecture of 	
	M-CNN is shown in Fig.~\ref{fig:cnn} and other variants are constructed by  gradually adding/eliminating  the cross image filters in different layers. 
	M-CNN contains $4$ cross image matching filters from layers $conv2$ to $conv5$. 
	``M-CNN (cross 5,4,3,2,1)''  is obtained by adding $11\times11\times6$ cross image filters in the ``conv1'' layer to the ``M-CNN''. 
	The added matching filters are applied on the stacked image composed  of the RGB channels of input image and KNN region. In addition, we continue to remove  the cross image matching filters layer by layer,  producing  ``M-CNN (cross 5,4,3)'', ``M-CNN (cross 5,4)'' , ``M-CNN (cross 5)'' and ``M-CNN(w/o cross)''. Note that  no matching filters are used in the  ``M-CNN(w/o cross)'' architecture, where ``w/o'' stands for without.
	For fair comparison, we keep the number of feature maps of each convolutinal layer unchanged for different M-CNN variations. 
	Therefore, the number of removed  cross image filters is evenly added to the corresponding two single image  layers. For example, the number of feature maps for both single and cross image convolutional paths are all $30$ in ``conv2''. After we remove the cross matching filters  to derive ``M-CNN (cross 5,4,3)'', ``M-CNN (cross 5,4)'' and ``M-CNN (cross 5)'', the number of feature maps in the single image convolutional path is set as $45$.
	In addition, we compare with a classic CNN based verification architecture named ``Siamese''~\cite{chopra2005learning} which is a composite structure of two identical sub-networks. The outputs of the two sub-networks are fully-connected linear layer responses, whose absolute differences are calculated as features to fit the final matching confidence and displacements. 
	Finally,  we compare with``{M-CNN w/o ss }''  which is the same with  M-CNN except that 
	the superpixel smoothing processing step is skipped. 
	
	The average scores over all  labels  in Table \ref{experiment_all_label} offer following observations.
	Incrementally adding cross image matching filters into more convolutional layers produces 4 variations of ``M-CNN'', including ``M-CNN (w/o cross)'', ``M-CNN (cross 5)'', ``M-CNN (cross 5,4)'', ``M-CNN (cross 5,4,3)'' and ``M-CNN''.
	Their $F_1$ scores increase from $56.99\%$, $58.07\%$, $60.36\%$ to $62.81\%$. 
	The highest $F_1$ score is $62.81\%$ which is reached by ``M-CNN''. 
	The gradually  improving performance validates that   inserting more cross image matching into multiple convolutional layers can help achieve better matching. 
	Adding a cross image matching kernel in the ``conv1'' layer drops the $F_1$ score from $62.81\%$ to $61.53\%$. 
	The reason for the relatively lower result is that the receptive field corresponding to  the first cross image matching kernel  is small and only involves certain part of a semantic label. However the targets of this work, i.e., the matching confidence and displacements, are defined on the semantic label level and thus are beyond the receptive field of the cross image matching kernel inserted in the ``conv1'' layer. 
	When the  M-CNN grows deeper, the receptive fields become much larger, with a higher probability to cover the semantic label, which can faciliate estimating the semantic label-level displacements.	In addition,   ``M-CNN (w/o cross)'' performs better than  ``Siamese'' because our ultimate task is to estimate  displacements.
	``M-CNN (w/o cross)'' calculates the difference between two ``conv5'' layer feature maps while ``Siamese'' calculates the differences of two fully connected features, where spatial structures in the $2$-dim images are partially lost. 
	The lower $F_1$  score of ``{M-CNN w/o ss}''  compared with ``M-CNN'' proves that the adopted superpixel smoothing technique can better preserve the 
	boundary information although it is a simple and fast voting  of pixels' labels. The superior performance of ``{M-CNN w/o ss}'' than 
	the state-of-the-arts~\cite{yamaguchi2012parsing,yamaguchi2013paper}  shows that our M-CNN has the capability of directly predicting more 
	reliable  label masks even without the post processing step.

	%
	%
	
\begin{figure}[h]
		\begin{center}
			\includegraphics[width=0.95\linewidth]{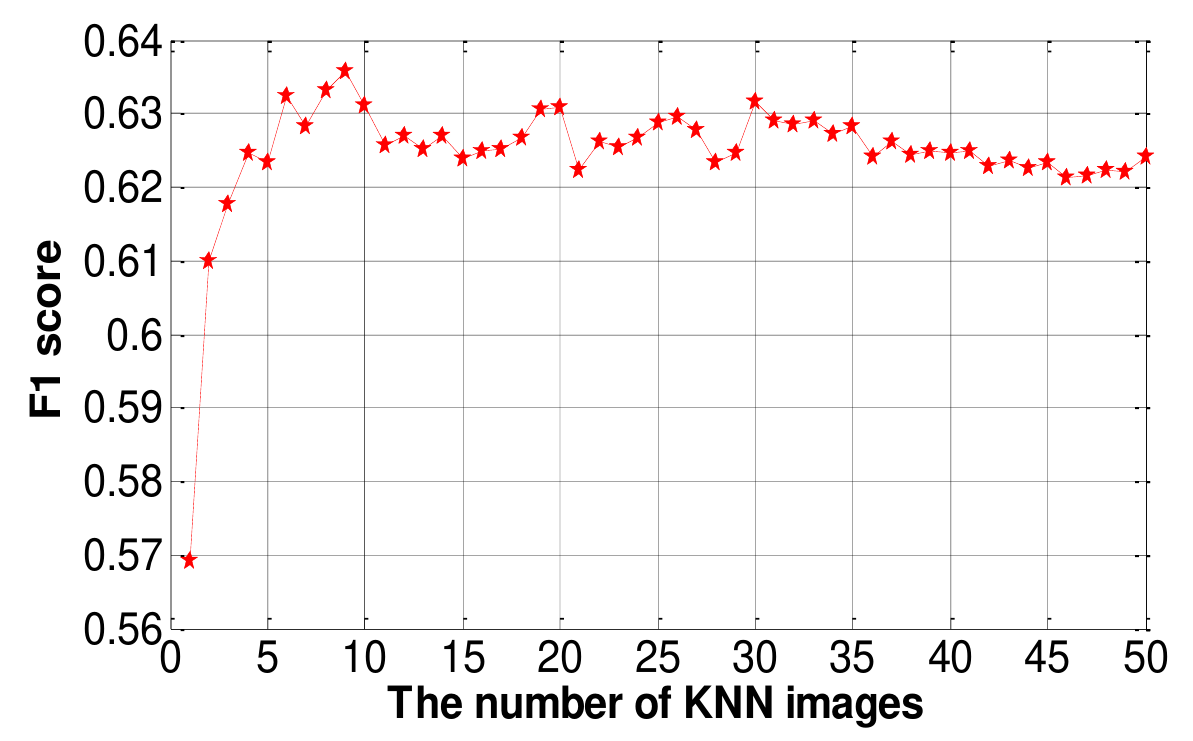}
		\end{center}
		\vspace{-0.1in}
		\caption{  $F_1$ scores for different numbers of KNN images.
		}
		\label{fig:knn_number}
		\vspace{-0.3in}
	\end{figure}

	\textbf{Sensitivity to the Number of $K$}:
	In Fig.~\ref{fig:knn_number}, we report the performance of our human parsing method with respect to different numbers of KNN images. 
	We find that M-CNN reaches the highest $F_1$ score $63.58\%$ when $9$  KNN regions are considered. 
	When only $1$ KNN  region is considered, the performance is still quite competitive ($56.92\%$).

	\textbf{Qualitative Parsing Results Comparison}:
	Fig. \ref{fig:final_fig} shows the comparison between M-CNN and  PaperDoll~\cite{yamaguchi2013paper}.  
	The results demonstrate that our method can successfully predict the label maps with small regions, which can be attributed to the reliable label transferring from the KNN regions. For example, the bag, scarf and hat in three images in the first column are successfully located by M-CNN but totally missed by PaperDoll.  Another example is that M-CNN successfully finds the small sunglasses in the first row, which are missed by PaperDoll.  In addition, it can be observed that the results from M-CNN are robust to pose variations. As shown in the bottom row, M-CNN can accurately estimate the locations of left and right arms, while PaperDoll cannot.  The superior performance is because our method is an end-to-end model while PaperDoll relies on a separate pose estimation preprocessing step.  Another observation of Fig. \ref{fig:final_fig} 	 is that  our segmented regions are more complete while the PaperDoll regions are fragmented, such as the lower right result. 
	That is because PaperDoll transfers labels based on oversegments which lack explicit  semantic meaning. 
	
	
			\vspace{-0.1in}
	
	\section{Conclusion and Future Work} \label{sec:conclusion}
	In this work, we tackle the human parsing problem by  proposing a new  quasi-parametric model. Our unified end-to-end quasi-parametric framework inherits the merits of both parametric and non-parametric parsing methodologies.
	It takes advantage of the supervision from annotated data and can be easily extended to newly annotated images and labels. To characterize the multi-ranged matching, we propose a Matching  Convolutional Neural Network, which  
	contains two single image convolutional paths for better feature representation and a cross image convolutional path where cross image matching filters are embedded into the convolutional layers.  
	Extensive experimental results clearly demonstrate
	significant performance gain from the quasi-parametric model over the state-of-the-arts.
	In the future, we will extend the framework  to other exemplar-based tasks, such as face parsing. 
	Moreover,  we plan to use other more power network structure, e.g, GoogLeNet~\cite{googleLeNet}.

	 \section*{Acknowledgement}
	 This work is supported by National Natural Science Foundation of China (No.61422213, 61332012, 61328205), and 100 Talents Programme of The Chinese Academy of Sciences.
	This work is  partly supported by gift funds from Adobe, 	National High-tech R$\&$D Program of China ($Y2W0012102$), 	 the Hi-Tech Research and Development Program of China (no.2013AA013801), Guangdong Natural Science Foundation (no.S2013050014548), and Program of Guangzhou Zhujiang Star of Science and Technology (no.2013J2200067).


	{
		\bibliographystyle{ieee}
		\bibliography{egbib}
	}

\end{document}